\pdfoutput=1
\documentclass[11pt]{article}

\usepackage{acl}
\usepackage{graphicx}
\usepackage{subfigure}
\usepackage{booktabs} % for professional tables
\usepackage{hyperref}
\usepackage{times}
\usepackage[T1]{fontenc}
\usepackage[utf8]{inputenc}

\usepackage{microtype}

\title{Interpretability of Fine-grained Classification \\ of Sadness and Depression}

\author{Tiasa Singha Roy*, Priyam Basu* \and Aman Priyanshu \\ 
Manipal Insitute of Technology \\
\{tiasa.singharoy, priyam.basu1, aman.priyanshu\}@learner.manipal.edu
\AND
Rakshit Naidu \\
Carnegie Mellon University \\
rakshitnaidu@cmu.edu}
  
\begin{document}
\maketitle
\begin{abstract}
While sadness is a human emotion that people experience at certain times throughout their lives, inflicting them with emotional disappointment and pain, depression is a longer term mental illness which impairs social, occupational, and other vital regions of functioning making it a much more serious issue and needs to be catered to at the earliest. NLP techniques can be utilized for the detection and subsequent diagnosis of these emotions. Most of the open sourced data on the web deal with sadness as a part of depression, as an emotion even though the difference in severity of both is huge. Thus, we create our own novel dataset illustrating the difference between the two. In this paper, we aim to highlight the difference between the two and highlight how interpretable our models are to distinctly label sadness and depression. Due to the sensitive nature of such information, privacy measures need to be taken for handling and training of such data. Hence, we also explore the effect of Federated Learning (FL) on contextualised language models. The code for this paper can be found at:~\footnote{\href{https://github.com/tiasa2/Interpretability-of-Federated-Learning-for-Fine-grained-Classification-of-Sadness-and-Depression}{https://github.com/tiasa2/Interpretability-of-Federated-Learning-for-Fine-grained-Classification-of-Sadness-and-Depression}}
\end{abstract}

\section{Introduction}
Mental Health is defined as a “state of well-being in which individuals realize their potential, cope with the normal stresses of life, work productively, and contribute to their communities” by The World Health Organization (WHO). Depression is a very common mental disease that a large number of people throughout the world suffer from. According to research conducted by the WHO, in 2020 more than 280 million people all over the world suffer from depression (WHO). It acts as leading cause for disability and is the most common form of neuro-psychiatric disorder~\cite{depressionwho}. According to Substance Abuse and Mental Health Services Association, in 2018, adolescents aged 12 to 17 years old had the highest rate of major depressive episodes (14.4\%) followed by young adults 18 to 25 years old (13.8\%). Older adults aged 50 and older had the lowest rate of major depressive episodes (4.5\%). Two-thirds of those who commit suicide struggle with depression~\cite{singlecare}.

% Sadness is an emotional state characterized by feelings of unhappiness and low mood. It is considered one of the basic human emotions. It is a normal response to situations that are upsetting, painful, or disappointing.  Other ways to talk about sadness might be ‘feeling low,’ ‘feeling down,’ or ‘feeling blue.’ A person may say they are feeling ‘depressed,’ but if it goes away on its own and doesn’t impact life in a big way, it probably isn’t the illness of depression~\cite{saddep,APA}. Unlike depression, which is persistent and longer-lasting, sadness is temporary and transitory~\cite{leonard_holmes_2021}. Hence, it is very important for us to be able to detect the difference between the two conditions as depression requires urgent care and treatment.

Sadness, on the other hand, is an emotional state characterized by feelings of unhappiness and low mood. A person may say they are feeling ‘depressed,’ but if it goes away on its own and does not impact life in a big way, it probably isn’t the illness of depression~\cite{saddep,APA}. Unlike depression, which is persistent and longer-lasting, sadness is temporary and transitory~\cite{leonard_holmes_2021}. Hence, it is very important for us to be able to detect the difference between the two conditions as depression requires urgent care and treatment.

\begin{figure*}[h]
\vskip 0.01in
\begin{center}
\centerline{\includegraphics[width=14cm]{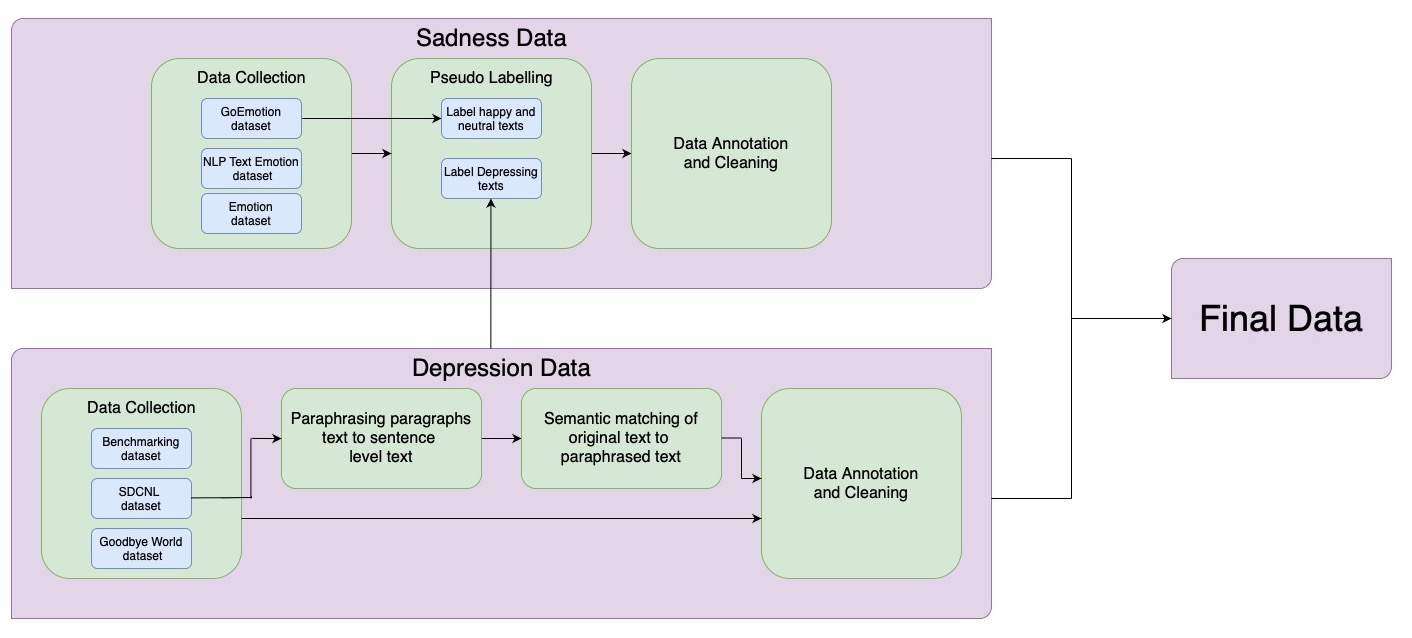}}
\caption{Pipeline of data creation system}
\label{fig:Figure1}
\end{center}
\vskip -0.01in
\end{figure*}

Most of the open-sourced data related to depression available also contain texts that imply just "sadness" as a part of it. Sadness related data, within itself, also comprises of depressing texts as seen in the GoEmotions dataset~\cite{demszky2020goemotions}. Sentences such as \emph{"I feel so much lonely"} and \emph{"I'm so depressed"} are found under the "sad" label. The converse, i.e sad texts are also often found under Depression corpora. In the Benchmarking dataset~\cite{basu2021benchmarking}, sentences like \emph{"Forgot my cheese cake at work"} and \emph{"Wish I had took holidays instead of being at work"}, which imply the narrator is sad but not depressed, are present under the "depression" label. Motivated by such problems, we propose a novel dataset where we try to tackle this problem by providing labelled samples of sad and depressing data separately to enable private training of machine learning models for detection of the presence of, as well as differentiation between, the two emotions. 
% [ENTER 2-3 LINES ABOUT EXPLORING INTERPRETABILITY AND EXPLAINABILITY AND WHY IT IS IMPORTANT FOR US.]
In this paper, we intend to explore how interpretable detecting sadness and depression are using common language models-- given that the definitions of both are very close to each other and yet, both are distinctly different. We also investigate whether Federated Learning (FL) could be a potential solution for training models on our sensitive dataset and if FL models lead to higher interpretable models than the baseline models as FL invloves collective aggregation of model updates.

\section{Related Works}
With the fast increase of social media usage, efforts for depression and other emotion detection are being made as a part of a Sentiment analysis task. ~\cite{park2015automatic} indicated that depressed Twitter users tend to post tweets containing negative emotional sentiments more than healthy users. In their paper~\cite{alsagri2020machine}, the authors explored multiple ML classification techniques for detecting depression on Twitter using content and activity features. Random forest techniques have also been used for detecting signs of sadness~\cite{cacheda2019early}. Deep learning approaches are popularly used for sentiment analysis task~\cite{babu2022sentiment} and have been explored later in this paper. However, none of these works target the contrast between sadness and depression. Hence, we create our own dataset which has been discussed in the next section. Alongside their powerful performance deep learning models also act as black-box prediction systems, therefore explainable methods become important for interpretable communication of model inference \cite{ribeiro2016why}. The ability of LIME a model-agnostic local method for interpretability introduction within text classification systems, has become well known for its impactful performance \cite{pmlr-v130-mardaoui21a}. We employ said XAI method for further substantiating our study.

\section{Sad-Depression Dataset Creation}

\subsection{Depression}
\subsubsection{Data Collection}
\begin{enumerate}
\item  SDCNL dataset: This paper~\cite{haque2021deep} was presented at ICANN 2021, proposing a dataset of depression vs suicide tweets, where the authors had scraped data from posts, that  belonged to the subreddits -  r/SuicideWatch and r/Depression and were labelled as suicidal and depressed respectively. Most of the data-points in a paragraph format having more than 2000 characters. We utilized the depression labelled posts in this dataset to create the initial set of depression labelled texts in our dataset. 
\item Goodbye World Dataset: This work aimed at identifying individuals at risk of suicide ~\cite{hesamuel}. The data collection here was similar to the one in the previous SDCNL dataset, using the Reddit API to collect posts from r/SuicideWatch and r/Depression. Unlike the SDCNL dataset, the data-points available here were mostly in a sentence or sentence-couplet format and thereby, better suited for the aim of our study. We used the posts from the depression labelled file which were non-identical from the ones we had earlier collected.
\item Benchmarking Dataset: This used as a dataset to benchmark transformer models~\cite{basu2021benchmarking}. It contains two labels - depression/sad(label 1) and non depression or neutral(label 0). For the purpose of this work, we utilized the data under the label depression. 

\end{enumerate}
\subsubsection{Paraphrasing}
As mentioned earlier, the data collected from SDCNL dataset contained majorly paragraphs, with more than 2000 characters in each, describing the depression context. Since our work aims to understand the distinction of context at a sentence level, we utilize a paraphrasing approach to get a sentence representation of the paragraph while retaining the equivalent contextual meaning, with the intention to retain the depression context. Google's Pegasus transformer model~\cite{zhang2020pegasus} was used for this task because of its abstractive summarization capabilities.

\subsubsection{Semantic Matching}
After paraphrasing, few of the paraphrased sentences were unable to capture the context of their respective paragraphs accurately. To get only the accurate sentences we utilized S-BERT~\cite{reimers2019sentence} to perform semantic matching between the paraphrased sentences and their original paragraphs. An arbitrary threshold value was set for selection of the correct sentences, which were then added to the data from SDCNL and goodbye world dataset to form the depression section of our dataset.

\subsubsection{Data Annotation and Cleaning}
For the final data under the "depression" label, we manually reviewed and annotated the processed data. It was carried out to remove ambiguous samples and to maintain the distinction between the two classes by re-labelling the samples that fit better under the "sadness" label. To differentiate between depression and sadness, the authors examined the context of the sentences in question based on the distinction made by American Psychiatric Association (APA)~\cite{APA}
After this cleaning was carried out on the data to remove special characters, external links and '@' tags. The cleaned data was labelled as 1 for depression. 

\subsection{Sadness}
\subsubsection{Data Collection}
\begin{enumerate}
\item GoEmotions Dataset: Created by Google Research~\cite{demszky2020goemotions}, GoEmotions is a human-annotated dataset of 58k Reddit comments extracted from popular English-language subreddits and labelled with 28 emotion categories. It includes 12 positive, 11 negative, 4 ambiguous emotion categories and 1 “neutral”. Within the negative, we picked comments labeled under sadness for the "sadness" label of our dataset.
\item NLP-text-emotion dataset: This dataset~\cite{lukasgarbas} was combined from dailydialog~\cite{li2017dailydialog}, isear~\cite{unige_2022}, and emotion-stimulus~\cite{ghazi2015detecting} to create a balanced dataset with 5 labels. The texts mainly consist of short messages and dialog utterances. We utilized it for the sadness related data for our dataset.
\item Emotions dataset for NLP: Open sourced dataset on Kaggle~\cite{praveen_2020}. This was created as a dataset for NLP classification tasks utilizing CARER methodology~\cite{saravia-etal-2018-carer} to generate 6 label categories. For our dataset, we used the data marked under the label - "sadness".
\end{enumerate}

\subsubsection{Pseudo-Labelling}
Upon inspection of the collected data we found an issue unique to the sadness labels. These labels contained, apart from sadness, sentences of neutral or even positive sentiment. To improve the quality of data under this label, we utilized a BERT~\cite{devlin2018bert} based classification model to assemble the correct ones, by training on the benchmarking dataset. We picked this dataset as it contained neutral or non depressing/sad samples(label 0) as well as depressing/sad samples(label 1). The trained model was then used to classify our collected "sadness" data to create pseudo labels to remove data points that are closer in sentiment to the "neutral" data and retaining the ones that have a closer relation to the depression/sad data. This allows us to essentially extract potentially "sad" data from the original collected samples. Finally all the sentences labelled as 1 are collected under the label sadness in our final dataset. 

\subsubsection{Data Annotation and Cleaning}
For the final data under the label "sadness", we again manually reviewed and annotated the processed data. Similar to the depression data, this was done to re-label any samples that are closer to depression rather than sadness or remove samples with ambiguous context. The same source was utilized as used for depression as a baseline definition while annotating in order to maintain uniformity.
We also removed a few neutral and positive context based samples that remained. To retain only "sad" samples, we focused on the definition provided by ~\cite{APA} to remove all the samples that did not align with the definition.
Finally cleaning was carried out on the data to remove special characters, external links and '@' tags. The cleaned data was labelled as 0 for sadness.

\subsection{Final Dataset}
At the end, we took all the depression-labelled and sadness-labelled data-points and combined them into a single dataset, which consisted of 3256 samples, of which 1914 samples were labelled as "sadness"(label 0) and 1342 samples under the label "depression"(label 1). It essentially follows a 0.58-0.42 class ratio, favouring effective training of classification models.

\section{Interpretability of Federated Models}

The rise in the use of machine learning models such as deep neural networks for intent mining has led to more accurate predictions. Explainable AI provides a set of methods to help us understand and interpret a model's decisions. In order to trust a model’s predictions, one must be able to interpret the reasons behind its decisions. Several attempts have been made to infer the text classification for language models. State of the art model-agnostic explanations such as LIME (Local Interpretable Model-Agnostic Explanations) have been used to enable better visualization and analysis of AI models.

% The rise in the use of machine learning models such as deep neural networks for intent mining has led to more accurate predictions.These algorithms behave like black boxes, making it difficult for practitioners to understand model predictions. Explainable AI provides a set of methods to help us understand and interpret a model's decisions. In order to trust a model’s predictions, one must be able to interpret the reasons behind its decisions. Several attempts have been made to infer the text classification for language models. State of the art model-agnostic explanations such as LIME (Local Interpretable Model-Agnostic Explanations) enable better visualization and analysis of AI models.

% LIME employs a local surrogate model, which means that it is a trained model used to approximate the predictions of the underlying black-box model. It utilizes variation generation of data to test the prediction of the target deep learning model. Using this perturbed data as a training set instead of using the original training data, it understands which tokens/words have the most impact on model classification prediction. LIME essentially trains an interpretable model, which is weighted by the proximity of the sampled instances to the predicted class.

\section{Experimental Results}

LIME is model agnostic in nature, that is, one can use it for any machine learning model. LIME allows interpretability to be supplemented to benchmark models. In our venture to integrate explainable AI for sadness-depression classification, we evaluated multiple models for performance comparison.

\subsection{Baseline Results}

\begin{figure*}[h]
\vskip 0.01in
\begin{center}
\centerline{\includegraphics[width=10cm]{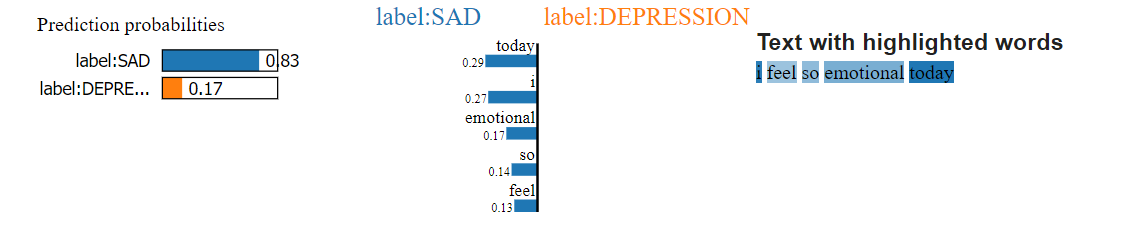}}
\caption{LIME base interpretation of Sentences---Class:SAD}
\label{fig:Figure2}
\end{center}
\vskip -0.01in
\end{figure*}

\begin{figure*}[h]
\vskip 0.01in
\begin{center}
\centerline{\includegraphics[width=12cm]{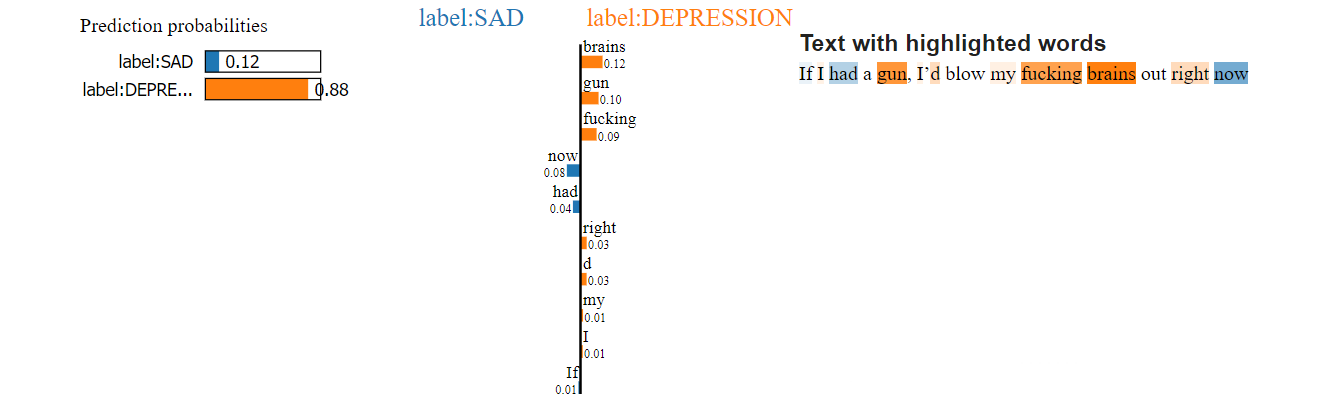}}
\caption{LIME base interpretation of Sentences---Class:DEPRESSION}
\label{fig:Figure3}
\end{center}
\vskip -0.01in
\end{figure*}

We intially conduct a detailed benchmark inference, over BERT and RoBERTa classification models. We provide these results in table~\ref{tab:table1}.

\begin{table}[h!]
    \caption{Performance of Baseline models}
    \label{tab:table1}
    \centering
    \begin{tabular}{|c|c|}
    \hline
         \textbf{Model Name} & \textbf{Accuracy} \\ 
         \hline
         BERT & 91.9\% \\
         \hline
         \textbf{RoBERTa} & \textbf{96.62\%} \\
        %  \hline
        %  DistilBERT & 92.93\% \\
        %  \hline
        %  ALBERT & 83.33\% \\
         \hline
    \end{tabular}
\end{table}

Inferring the performance presented in Table \ref{tab:table1}, we can see that the RoBERTa model is the best performing model and acts as the appropriate candidate for further exploration using explainable AI. Alongside RoBERTa we also take a look at the performance of BERT as a control for our analysis.

\subsection{Federated Setting Results}

We further reproduce these results in a federated setting. For this we employ the federated averaging algorithm, we present these results in table~\ref{tab:table2}. The table discusses results in both an IID-setting and a Non-IID setting, allowing us to substantiate the use of federated learning for privacy-enabled classification of this sensitive data.

\begin{table}[h!]
    \caption{Performance in Federated Setting}
    \label{tab:table2}
    \centering
    \begin{tabular}{|c|c|c|}
    \hline
         \textbf{Model Name} & \textbf{IID Setting} & \textbf{Non-IID Setting} \\ 
         \hline
         BERT & 75.07\% & 73.6\% \\
         \hline
         \textbf{RoBERTa} & \textbf{81.85\%} & \textbf{83.66\%} \\
         \hline
    \end{tabular}
\end{table}

For our experiments, we synthetically simulate $K$---Total number of clients clients, with $C$---Fraction of clients randomly chosen to perform computation on each round, and $E$---Number of training passes each client makes over its local dataset per round. For our inferential setting, we set $K=10, C=0.3$, and $E=1$ for the experiments. Having produced successful convergence, with satisfactory performance, we can claim the utility of federated learning and explainable AI for sadness---depression classification.

We also demonstrate the utility and inference that LIME brings to this federated system. In a study on topics as delicate as depression classification, it becomes integral that over-seeing authorities/deployers understand the inferential steps the model may be taking. We supplement samples of explainability for Sadness and Depression detection in Figure~\ref{fig:Figure2} and Figure~\ref{fig:Figure3}. Figure~\ref{fig:Figure2} deliberates on the case of Sadness detection for the sentence "I feel so emotional today," where the LIME interprets model decision on perturbed word analysis. As one can see, the impact of the words "today" and "i" hold significant weightage to model inference for class label 0 or SAD. On the other hand, Figure~\ref{fig:Figure3} deliberates on the case of Depression detection for the sentence "If I had a gun, I'd blow my fu*king brains out right now," where LIME distinctly highlights the words "brains", "gun", and "fu*king" which significantly impact the model's prediction to be class label 1 or DEPRESSION. We can ascertain that certain words like "now," "had," and "if" had an opposing effect; however, their impact wasn't as consequential as that of the aforementioned highlighted words. Employing interpretability, therefore, allows more accessible and successful affirmation of model predictions during deployment even within a federated setting, thereby substantiating the experimental setting.

\section{Conclusion}

Sadness and depression both are sensitive attributes of one's life. Social media has given a new platform for people to vent their feelings, at the same time, an opportunity for one to detect impactful signs early on. The application of SOTA interpretable methods on this assortment of data will pave a path towards the adoption of the same during real-world implementation. Our use of LIME, which quantify local model explanations, allows us to display their importance and relevance in depression sadness classification. For future explorations, we would like to explore these explainable techniques for the same. We would also like to investigate other techniques such as lexical normalization to reduce noise and measure its impact on our predictions. We believe that our work serves as a valuable resource for sadness and depression classification. The integration with explainable as well as privacy-preserving methods paves a new path towards future research and development.

\bibliography{main}

\end{document}